%% file: main.tex
\documentclass[letterpaper, 10 pt, conference]{ieeeconf}

\IEEEoverridecommandlockouts                              

\makeatletter
\let\NAT@parse\undefined
\makeatother
\usepackage[backref=page]{hyperref}
\hypersetup{colorlinks=true, 
	unicode=true, 
	linkcolor=[rgb]{0.10,0.05,0.67}, 
	citecolor=green, 
	filecolor=[rgb]{0.10,0.05,0.67}, 
	urlcolor=[RGB]{237, 15, 152} 
}

\usepackage{graphics} 
\usepackage{epsfig} 
\usepackage{mathptmx} 
\usepackage{times} 
\usepackage{amsmath} 
\usepackage{amssymb}  
\usepackage{url}
\usepackage{booktabs}
\usepackage{siunitx}
\usepackage{csquotes}
\usepackage{pgf}
\usepackage{pgfplots}
\usepackage[T1]{fontenc}

\usepackage{multirow} 
\usepackage{pifont} 
\usepackage{arydshln} 
\usepackage{subfiles}
\usepackage{balance} 
\usepackage[absolute]{textpos} 
\usepackage{xcolor}
\usepackage{graphicx} 
\usepackage[skip=2pt]{subcaption} 

\usepackage{pifont}
\newcolumntype{x}[1]{>{\centering\arraybackslash\hspace{0pt}}p{#1}}

\usepackage[skip=2pt]{caption}

\usepackage{etoolbox}
\makeatletter
\patchcmd{\@makecaption}
  {\scshape}
  {}
  {}
  {}
\makeatletter
\patchcmd{\@makecaption}
  {\\}
  {.\ }
  {}
  {}
\makeatother

\title{Diffusion-Based Image Augmentation \\ for Semantic Segmentation in Outdoor Robotics}

\input{tex/teaser_figure.tex}

\author{Peter Mortimer and Mirko Maehlisch\\
Perception for Autonomous Driving Lab\\
University of the Bundeswehr Munich\\
peter.mortimer@unibw.de
\thanks{The authors gratefully acknowledge funding by the Federal Office of
Bundeswehr Equipment, Information Technology and In-Service Support
(BAAINBw).}}

\usepackage{fancyhdr}
\fancypagestyle{withfooter}{
  
  \fancyfoot[C]{\footnotesize Presented at the 2025 IEEE ICRA Workshop on Field Robotics}
}

\begin{document}
\input{tex/rainbow_colors}
\input{tex/goose_colors}
\maketitle

\thispagestyle{withfooter}
\pagestyle{withfooter}

\begin{abstract}
The performance of leaning-based perception algorithms suffer when deployed in out-of-distribution and underrepresented environments.
Outdoor robots are particularly susceptible to rapid changes in visual scene appearance due to dynamic lighting, seasonality and weather effects that lead to scenes underrepresented in the training data of the learning-based perception system.
In this conceptual paper, we focus on preparing our autonomous vehicle for deployment in snow-filled environments.
We propose a novel method for diffusion-based image augmentation to more closely represent the deployment environment in our training data.
Diffusion-based image augmentations rely on the public availability of vision foundation models learned on internet-scale datasets. 

The diffusion-based image augmentations allow us to take control over the semantic distribution of the ground surfaces in the training data and to fine-tune our model for its deployment environment. 
We employ open vocabulary semantic segmentation models to filter out augmentation candidates that contain hallucinations.

We believe that diffusion-based image augmentations can be extended to many other environments apart from snow surfaces, like sandy environments and volcanic terrains.
\end{abstract}

\IEEEpeerreviewmaketitle

\section{Introduction}

The size and annotation granularity of semantic segmentation datasets for outdoor robotics is steadily increasing~\cite{mortimer_outdoorsurvey_2024}. 
The necessity of providing multi-season outdoor datasets is starting to be addressed. 
Datasets like ROVER~\cite{schmidt_rover_2024}, UTIAS Long-Term~\cite{mactavish_utias_2018} and FoMo~\cite{boxan_fomo_2024} focus on Visual SLAM across multiple seasons.
For 2D image and 3D LiDAR semantic segmentation, datasets like GOOSE~\cite{mortimer_goose_2024} and GOOSE-Ex~\cite{pemo:hagmanns_goose-ex_2025} contain annotations of scenes across all seasons.
We augment the outdoor dataset data from GOOSE with state-of-the-art image synthesis methods.
The emergence of diffusion probabilistic models, that model the image synthesis process as a sequential application of denoising autoencoders~\cite{ho_diffusion_2020, dickstein_diffusion_thermodynamics_2015}, has led to algorithms that outperform existing approaches based on Generative Adversarial Networks~\cite{dhariwal_diffusion_vs_gan_2021}.
This approach was improved in the form of latent diffusion models~\cite{rombach_sd2_2022}, also referred to as stable diffusion models, that enable text-to-image and image-to-image generation with possibilities to constrain and guide the generation process~\cite{zhang_controlnet_2023}.
Diffusion models are pretrained on internet-scale datasets like LAION~\cite{schuhmann_laion5b_2022}, giving them a good understanding of seasonal changes in natural images. 
\begin{figure}[htpb]
  \centering
  \resizebox{.8\linewidth}{!}
{\input{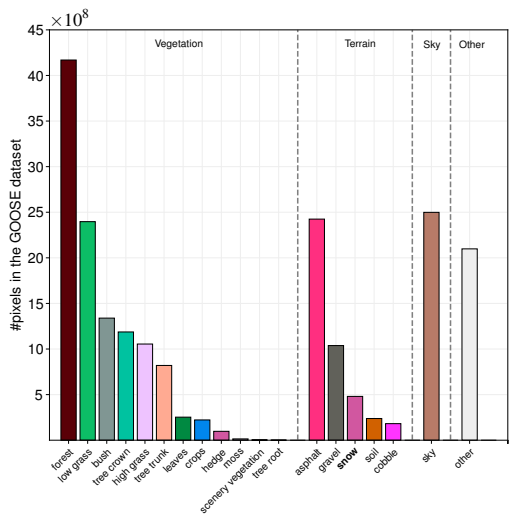}}
  \caption{Best to inspect digitally. Histogram of the annotated pixels in the 2D images of the GOOSE dataset. The 3 categories $\{$\textit{Vegetation}, \textit{Terrain}, \textit{Sky}$\}$ make up $89.7\%$ of the annotated pixels. The remaining categories are accumulated in \textit{Other}.
  Notice the small amount of annotated \textit{snow}~\textcolor{snow}{$\blacksquare$} pixels ($2.3\%$ of all annotated pixels in the GOOSE dataset) in comparison to more common ground surface classes like \textit{low grass}~\textcolor{low_grass}{$\blacksquare$}, \textit{asphalt}~\textcolor{asphalt}{$\blacksquare$} and \textit{gravel}~\textcolor{gravel}{$\blacksquare$}.}
  \label{fig:2d_histogram}
\end{figure}

\section{Related Work}

\subsection{Augmentations for Semantic Image Segmentation}

The use of affine transformations on input images as augmentation method for a learning-based system appear in early work on handwritten digit recognition~\cite{lecun_digit_recognition_1995}. 
The first convolutional neural networks of the deep learning era that were trained for semantic segmentation tasks also mention augmentations like image scaling, color jitter, horizontal flipping and image rotations~\cite{farabet_semseg_2013, long_fcn_2015}. In contrast to image classification tasks, the ground truth semantic mask has to be augmented in the same manner to preserve the consistency of the semantic pixel mapping.
The data augmentation methods for image classification have extended to operations that greatly affect image content beyond recognition like mixup~\cite{zhang_mixup_2018}, Cutout~\cite{devries_cutout_2017} and CutMix~\cite{yun_cutmix_2019}, with Copy-Paste~\cite{ghiasi_copypaste_2021} resembling the semantic segmentation extension of this trend.
Methods like Moment Exchange~\cite{li_feature_normalization_2021} augment data directly in feature space.
Rigoll et al. used CycleGAN~\cite{zhu_cycleGAN_2017} with domain knowledge to place traffic signs in semantically valid position in camera images as augmentation to improve the traffic sign detection~\cite{rigoll_traffic_sign_placement_2022}.
Our proposed data augmentation method also relies on learning-based image manipulation (diffusion-based in our case) with domain knowledge specific to outdoor robotics.

\subsection{Domain Adaptation for Semantic Segmentation}

For unsupervised domain adaptation from synthetic to real-world images, DAFormer~\cite{hoyer_daformer_2022} with its transformer-based architecture, has shown greater improvements than previous CNN-based architectures~\cite{tsai_adaptsegnet_2018}.
Test-time domain adaptation methods like CoTTA lay the focus on adapting the neural network while encountering the continually changing target environment~\cite{wang_cotta_2022}.
We assume for our approach, that we are given enough time to prepare our neural network before deployment in the underrepresented target domain.
The most similar work is DIDEX~\cite{niemeijer_didex_2024} which generalizes the trained source domain with diffusion-based augmentations generated by text prompts. 
Our approach relies on constraining the image synthesis process to such an extent, that the original semantic maps can easily be adapted to the augmented images.

\subsection{Winter Outdoor Robotics}

For urban autonomous driving, unlabeled research datasets like CADC~\cite{pitropov_cadc_2021} and Boreas~\cite{burnett_boreas_2023} were collected to evaluate perception pipelines in snow-filled driving conditions.
The recorded LiDAR scans of WADS~\cite{kurup_wads_2021} are semantically annotated in the common SemanticKITTI~\cite{behley_semantickitti_2019} format.
Additionally, WADS introduces semantic classes for accumulated snow and falling snow to train learning-based methods to handle the false positives generated from reflections on falling snowflakes.
The recent FinnWoodlands~\cite{lagos_finn_2023} includes semantically segmented annotations of forest scenes in winter. 
The overall focus in FinnWoodlands is on forestry-specific tasks like the panoptic segmentation of tree trunks for the classification of different types of trees.
GOOSE~\cite{mortimer_goose_2024} contains semantically segmented images from outdoor scenes across all four seasons. Due to snow only appearing in the winter recordings, this amounts to 14\% of all recorded images containing the semantic class \textit{snow} (see Figure~\ref{fig:2d_histogram}).
To adapt a learning-based perception system for deployment in a snow-filled environment, more training data is required.

\begin{figure*}[h]
    \centering
    \includegraphics[width=\linewidth]{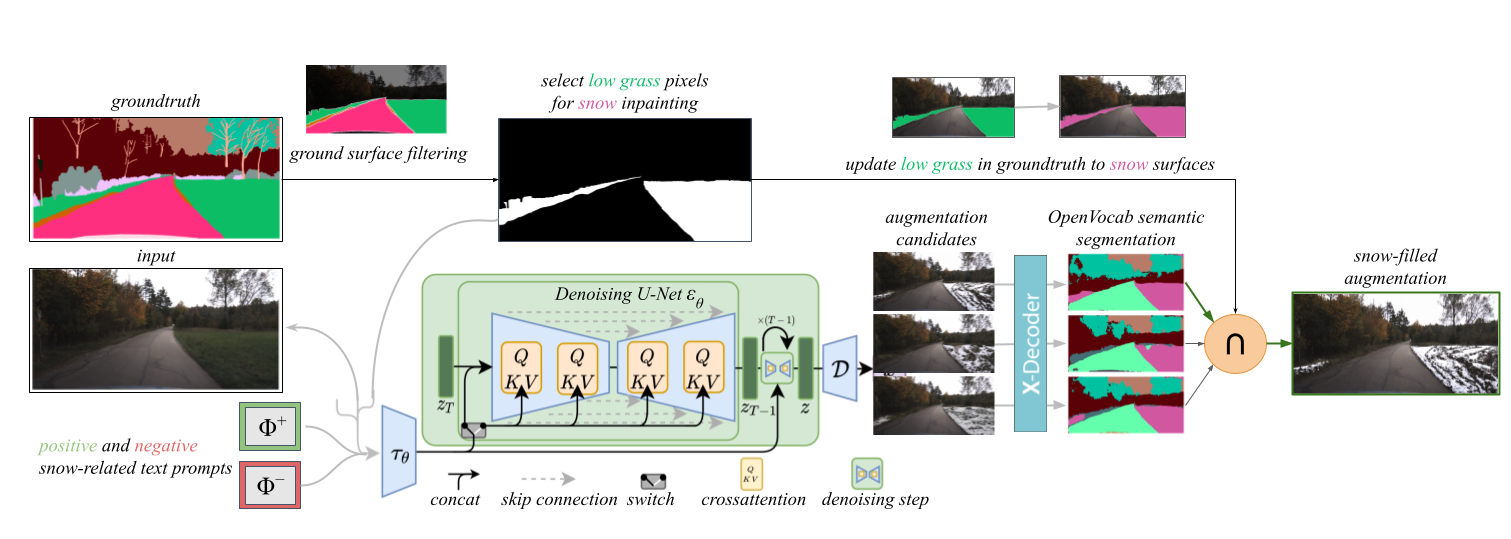}
    \caption{An overview of the diffusion-based image augmentation method. The diffusion-based image synthesis is conditioned with the original training sample as initial image, the in-painting mask selected from a subset available ground surfaces in the groundtruth image and the constant positive and negative text prompts $\Phi^{+}$ and $\Phi^{-}$. The denoising network is based on stable diffusion 2~\cite{rombach_sd2_2022} with an additional training step for the in-painting capability~\cite{suvorov_lama_2022}. The second stage uses X-Decoder~\cite{zou_xdecoder_2023} for open-vocabulary semantic segmentation to remove augmentation candidates with hallucinated obstacles and to select the augmentation candidate with the highest area of overlap to the expected groundtruth mask.}
    \label{fig:method_overview}
\end{figure*}

\section{Methodology}

\subsection{Diffusion-Based Image Augmentation}

The goal of diffusion-based image augmentation is to increase the robustness of the semantic segmentation in snow-filled environments.
We believe that increasing the appearance of snow in the training images will lead to improvements in the semantic segmentation when evaluated in snow-filled environments.
Our analysis is done on the GOOSE dataset, which stands out by containing semantically segmented scenes across all four seasons.
And while snow is not uncommon in the winter season, only 14\% of the images contain the semantic class \textit{snow}.
Viewed on the pixel-level, only 2.3\% of the annotated pixels are of class \textit{snow} (see Figure~\ref{fig:2d_histogram}).
With categories in natural images following a Zipfian distribution~\cite{gupta_lvis_2019}, we observe the same in the GOOSE dataset where many categories contain only few training samples.
Instead of making changes in the network architecture like region re-balancing for rare classes~\cite{cui_region_rebalancing_2022}, we approach the problem by changing the overall class distribution in the dataset.

The components that make up the diffusion-based image augmentation process are displayed in Figure~\ref{fig:method_overview} and can be divided into two stages, the image synthesis and the hallucination filtering. 

\noindent\textbf{Image Synthesis:} For the image synthesis we use the stable diffusion 2 model that was additionally trained using the mask-generation strategy from LaMa~\cite{suvorov_lama_2022}. This constrains the diffusion process to only in-paint the input image in the selected binary mask.
Since we have a good understanding of the semantic meaning of the pixels in semantic segmentation datasets like GOOSE, we can use the groundtruth semantic mask of a training image to select the areas that should be in-painted during the diffusion process.
Here we select a random subset of the ground surfaces present in the training image to generate the in-painting mask.
The conditioning that primarily drives the denoising network towards snow-filled surfaces is the positive textual prompt $\Phi^{+}$. 
The text prompt is encoded into the latent space that can be passed to the immediate layers of the denoising U-Net via cross-attention.
Similar approaches like DIDEX~\cite{niemeijer_didex_2024} used the class names of the groundtruth semantic mask as sub-strings of the positive text prompt to emphasize the semantic content expected in the denoised image (e.g.: $\Phi^{+}$ = \textit{"An image containing gravel, low grass, forest, snow,..."}).
We could not observe any advantages using this approach during our experiments with the diffusion-based image synthesis. We obtained our best results by using the following fixed text prompt: 
\begin{align*}
    \Phi^{+} = \textit{"A high quality photo; Covered in white snow."}
\end{align*}
The contrasting concept to the positive text prompt is the encoding of a negative text prompt $\Phi^{-}$ into the latent space with concepts that should not be in-painted. For this we used:
\begin{align*}
    \Phi^{-} = \textit{"Blurry parts, fences, any other obstacles, } \\
    \textit{visible grass patches, humans, dogs,} \\
    \textit{any faces, pedestrians, rocks, boulders."}
\end{align*}
Giving the diffusion process too few input conditions leads to hallucination artifacts like human limbs or rock boulders that are added onto the snow surface.
The negative text prompt $\Phi^{-}$ helped reduce the share of hallucination artifacts in the denoised images.

For the diffusion process, we observed sufficient change in the in-painted area and a convergence between subsequent denoising step after roughly 20 diffusion and denoising steps.  
The runtime of the diffusion process for a single mini-batch can take up to one minute on a NVIDIA RTX 4000 Ada.

\begin{figure*}[htpb]
    \centering
    \includegraphics[width=\linewidth]{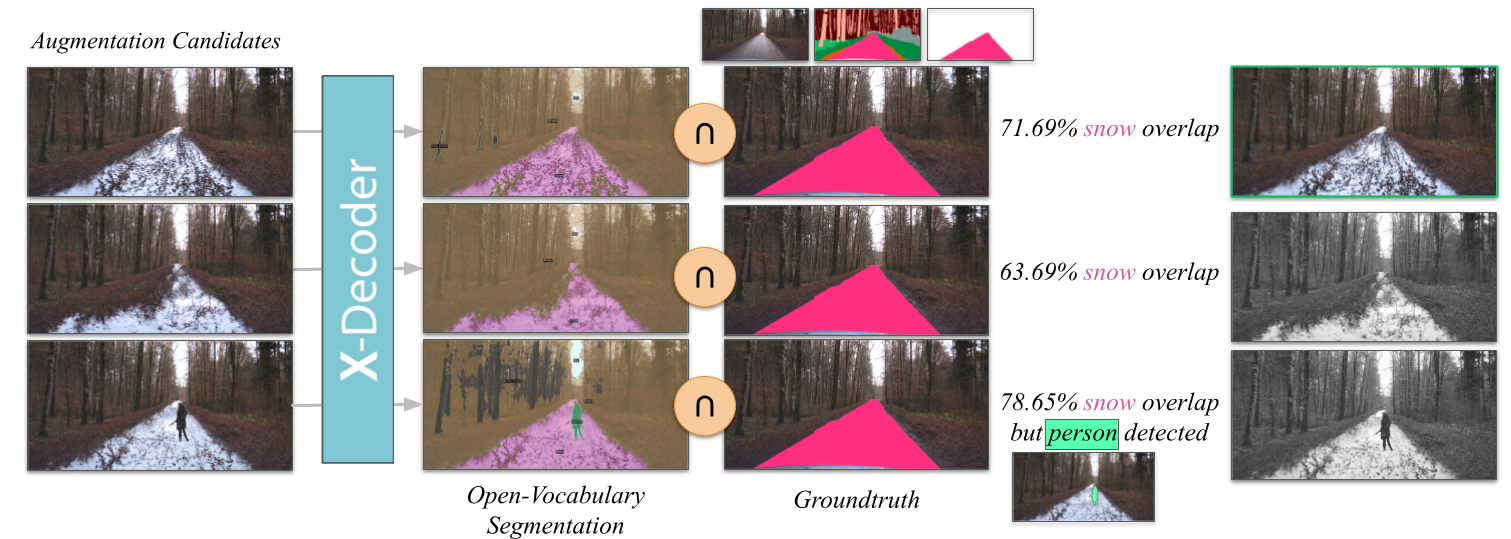}
    \caption{We can generate multiple augmentation candidates from the same input image by changing the seed.
    The diffusion model is prone to hallucinating people, bushes and animals onto the in-painting surface. An open-vocabulary segmentation model like X-Decoder~\cite{zou_xdecoder_2023} can reliably detect the objects. 
    We then select the candidate with the highest snow overlap that doesn't contain any hallucinations.}
    \label{fig:xdecoder_person}
\end{figure*}
\begin{figure}[h]
    \centering
    \includegraphics[width=\linewidth]{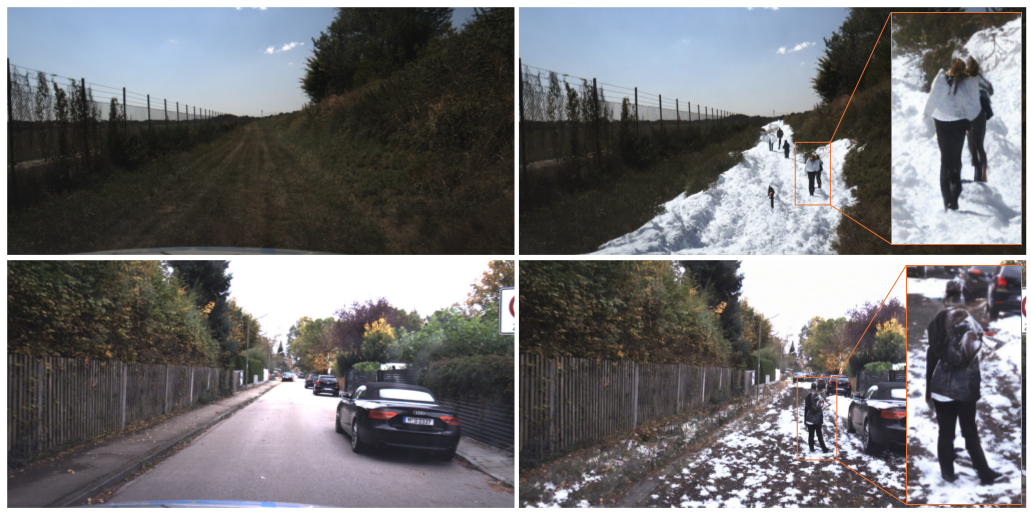}
    \caption{The in-painting process can lead to hallucinations by the diffusion model. Here are two examples where people-like artifacts were added during the diffusion process. The process visualized in Figure~\ref{fig:xdecoder_person} filters out these hallucinations.}
    \label{fig:hallucation}
\vspace{-2.em}
\end{figure}
\noindent\textbf{Hallucination Filtering:} 
Depending on the starting random seed, you obtain multiple potential augmentation candidates for the same set of textual and visual input conditions. 
At this stage, the generated images can still vary in quality with some containing hallucinations of objects on the snow-filled environment (see Figure~\ref{fig:hallucation}).
To be able to discern the semantic content in each augmentation candidate, we use the open-vocabulary segmentation of X-Decoder~\cite{zou_xdecoder_2023}, where we pass the labels of the GOOSE dataset as vocabulary.
This allows us to detect hallucinated obstacles like pedestrians on the snow (see Figure~\ref{fig:xdecoder_person}).
We determine the best augmentation candidates for a given scene by comparing the expected groundtruth mask with the semantic segmentation from X-Decoder.
The expected groundtruth mask consists of the GOOSE groundtruth mask for the input scene with the pixel areas selected for in-painting changed to our target ground surface class \textit{snow}.
Any augmentation candidate that contains hallucinated obstacles like pedestrians or fences in the in-painted area are discarded. 
Of the remaining augmentation candidates, the candidate with the highest area of overlap to the expected groundtruth mask is selected.

\begin{figure}[htpb!]
    \centering
    \includegraphics[width=.8\linewidth]{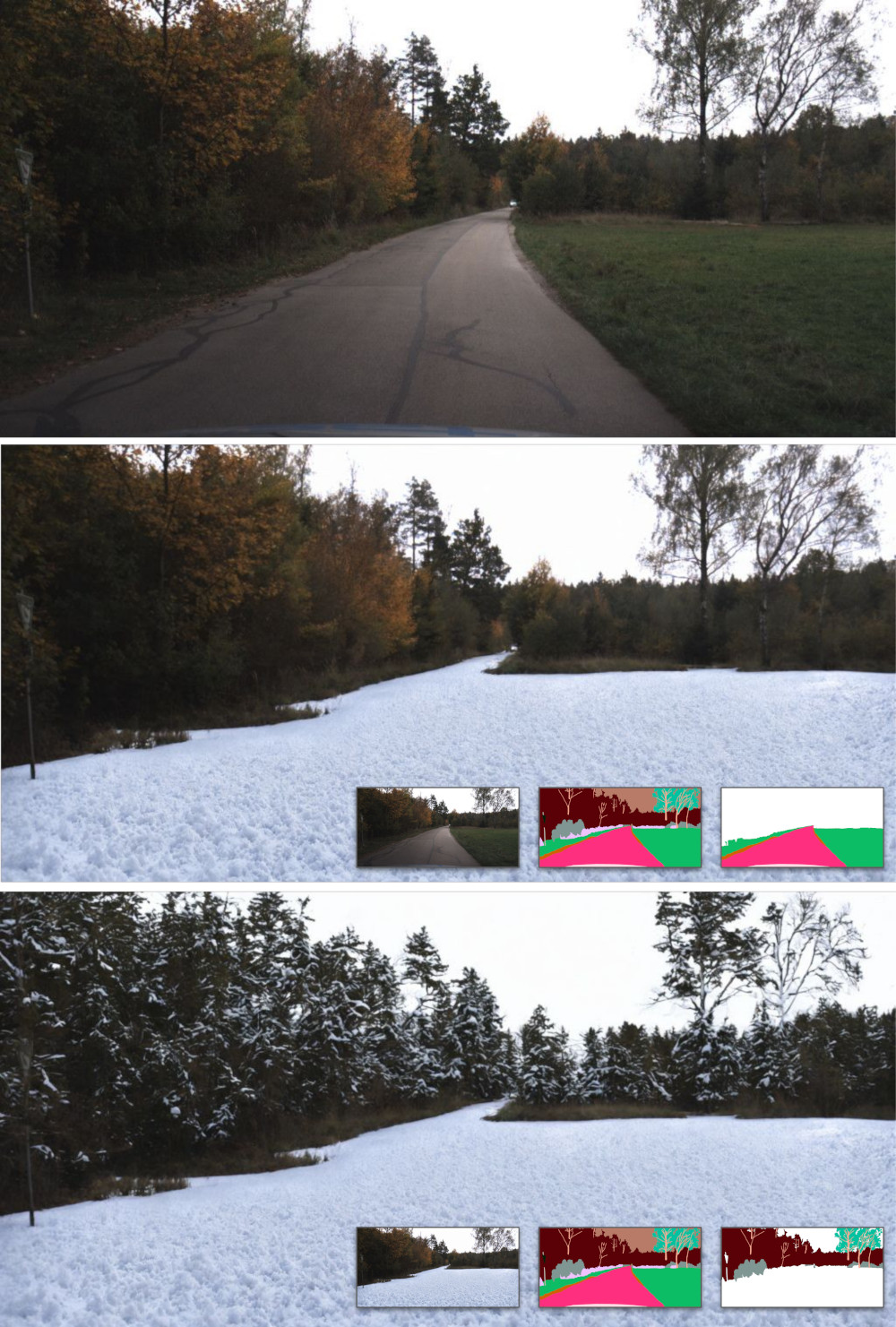}
    \caption{Our presented method lays a focus on changing the original ground surface (top) to a snow-filled surface (middle), but the image synthesis can also be applied to add a wintry appearance to the surrounding landscape (bottom). Here we obtained the best results by constraining the diffusion process for the landscape in-painting to 15 steps. The three inset images in the bottom-right display the input image for the diffusion process, the full semantic mask as reference and the in-painting mask used as input condition for the diffusion process.}
    \label{fig:inpainting_landscape}
\end{figure}

\section{Outlook}

We present a novel data augmentation method that can increase the occurrence of rare surface types like \textit{snow} in a training set by leveraging foundation models for image synthesis and open-vocabulary semantic segmentation.
This initial concept lacks a quantitative analysis on the best transfer learning scheme to improve the semantic segmentation of \textit{snow} from a model originally trained on a multi-season dataset. We also plan an analysis on the amount of augmented samples required for noticeable improvements in the snow surface segmentation.
This is kept as future work that builds on the diffusion-based image augmentation.
We also see potential in extending the augmentation process to in-painting wintry features to the surrounding landscape in the images~(see Figure~\ref{fig:inpainting_landscape}). 

\balance
\bibliographystyle{IEEEtran}
\bibliography{bib/IEEEfull,bib/additional_full,bib/literature}
\end{document}

%% file: tex/teaser_figure.tex
\let\oldtwocolumn\twocolumn
\renewcommand\twocolumn[1][]{%
	\oldtwocolumn[{#1}{
		\input{tex/rainbow_colors}
        \begin{center}
        \hspace*{-0.175\linewidth} 
        \resizebox{1.3\linewidth}{!}{\input{images/teaser_gallery/teaser_grid.pgf}}
        {
        Fig. 1.~Diffusion models allow for complex image augmentations like adding snow surfaces in annotated scenes from the GOOSE dataset~\cite{mortimer_goose_2024}. The images in the \textcolor[RGB]{233, 113, 50}{\textbf{orange frame}} are augmentated versions of the image in the previous column.
        \label{fig:teaser}
        }
		\end{center}
	}]
}

%% file: tex/rainbow_colors.tex
\definecolor{color1}{RGB}{255, 0, 0}
\definecolor{color2}{RGB}{255, 38, 0}
\definecolor{color3}{RGB}{255, 75, 0}
\definecolor{color4}{RGB}{255, 113, 0}
\definecolor{color5}{RGB}{255, 150, 0}
\definecolor{color6}{RGB}{255, 188, 0}
\definecolor{color7}{RGB}{255, 225, 0}
\definecolor{color8}{RGB}{239, 247, 0}
\definecolor{color9}{RGB}{207, 254, 0}
\definecolor{color10}{RGB}{170, 255, 0}
\definecolor{color11}{RGB}{133, 255, 0}
\definecolor{color12}{RGB}{95, 255, 0}
\definecolor{color13}{RGB}{58, 255, 0}
\definecolor{color14}{RGB}{20, 255, 0}
\definecolor{color15}{RGB}{8, 255, 25}
\definecolor{color16}{RGB}{0, 255, 54}
\definecolor{color17}{RGB}{0, 255, 92}
\definecolor{color18}{RGB}{0, 255, 130}
\definecolor{color19}{RGB}{0, 255, 168}
\definecolor{color20}{RGB}{0, 255, 205}
\definecolor{color21}{RGB}{0, 245, 234}
\definecolor{color22}{RGB}{0, 226, 253}
\definecolor{color23}{RGB}{0, 191, 255}
\definecolor{color24}{RGB}{0, 154, 255}
\definecolor{color25}{RGB}{0, 116, 255}
\definecolor{color26}{RGB}{0, 78, 255}
\definecolor{color27}{RGB}{0, 40, 255}
\definecolor{color28}{RGB}{12, 15, 255}
\definecolor{color29}{RGB}{34, 0, 255}
\definecolor{color30}{RGB}{72, 0, 255}
\definecolor{color31}{RGB}{109, 0, 255}
\definecolor{color32}{RGB}{147, 0, 255}
\definecolor{color33}{RGB}{185, 0, 255}
\definecolor{color34}{RGB}{220, 0, 252}
\definecolor{color35}{RGB}{251, 0, 245}
\definecolor{color36}{RGB}{255, 0, 211}
\definecolor{color37}{RGB}{255, 0, 174}
\definecolor{color38}{RGB}{255, 0, 136}
\definecolor{color39}{RGB}{255, 0, 99}
\definecolor{color40}{RGB}{255, 0, 61}

%% file: images/teaser_gallery/teaser_grid.pgf
\begingroup%
\makeatletter%
\begin{pgfpicture}%
\pgfpathrectangle{\pgfqpoint{0.000in}{0.5in}}{\pgfqpoint{9.830400in}{5.30000in}}%
\pgfusepath{use as bounding box, clip}%
\begin{pgfscope}%
\pgfsetbuttcap%
\pgfsetmiterjoin%
\pgfsetlinewidth{0.000000pt}%
\definecolor{currentstroke}{rgb}{1.000000,1.000000,1.000000}%
\pgfsetstrokecolor{currentstroke}%
\pgfsetstrokeopacity{0.000000}%
\pgfsetdash{}{0pt}%
\pgfpathmoveto{\pgfqpoint{0.000000in}{0.000000in}}%
\pgfpathlineto{\pgfqpoint{9.830400in}{0.000000in}}%
\pgfpathlineto{\pgfqpoint{9.830400in}{6.000000in}}%
\pgfpathlineto{\pgfqpoint{0.000000in}{6.000000in}}%
\pgfpathlineto{\pgfqpoint{0.000000in}{0.000000in}}%
\pgfpathclose%
\pgfusepath{}%
\end{pgfscope}%
\begin{pgfscope}%
\pgfpathrectangle{\pgfqpoint{1.274985in}{4.408846in}}{\pgfqpoint{1.784123in}{0.871154in}}%
\pgfusepath{clip}%
\pgfsys@transformshift{1.274985in}{4.408846in}%
\pgftext[left,bottom]{\includegraphics[interpolate=true,width=1.791667in,height=0.875000in]{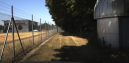}}%
\end{pgfscope}%
\begin{pgfscope}%
\pgfpathrectangle{\pgfqpoint{3.189007in}{4.408846in}}{\pgfqpoint{1.784123in}{0.871154in}}%
\pgfusepath{clip}%
\pgfsys@transformshift{3.189007in}{4.408846in}%
\pgftext[left,bottom]{\includegraphics[interpolate=true,width=1.791667in,height=0.875000in]{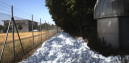}}%
\end{pgfscope}%
\begin{pgfscope}%
\pgfpathrectangle{\pgfqpoint{3.189007in}{4.408846in}}{\pgfqpoint{1.784123in}{0.871154in}}%
\pgfusepath{clip}%
\pgfsetbuttcap%
\pgfsetmiterjoin%
\pgfsetlinewidth{3.011250pt}%
\definecolor{currentstroke}{rgb}{0.913725,0.443137,0.196078}%
\pgfsetstrokecolor{currentstroke}%
\pgfsetdash{}{0pt}%
\pgfpathmoveto{\pgfqpoint{3.189007in}{4.408846in}}%
\pgfpathlineto{\pgfqpoint{4.973130in}{4.408846in}}%
\pgfpathlineto{\pgfqpoint{4.973130in}{5.280000in}}%
\pgfpathlineto{\pgfqpoint{3.189007in}{5.280000in}}%
\pgfpathlineto{\pgfqpoint{3.189007in}{4.408846in}}%
\pgfpathclose%
\pgfusepath{stroke}%
\end{pgfscope}%
\begin{pgfscope}%
\pgfpathrectangle{\pgfqpoint{5.103030in}{4.408846in}}{\pgfqpoint{1.784123in}{0.871154in}}%
\pgfusepath{clip}%
\pgfsys@transformshift{5.103030in}{4.408846in}%
\pgftext[left,bottom]{\includegraphics[interpolate=true,width=1.791667in,height=0.875000in]{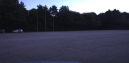}}%
\end{pgfscope}%
\begin{pgfscope}%
\pgfpathrectangle{\pgfqpoint{7.017052in}{4.408846in}}{\pgfqpoint{1.784123in}{0.871154in}}%
\pgfusepath{clip}%
\pgfsys@transformshift{7.017052in}{4.408846in}%
\pgftext[left,bottom]{\includegraphics[interpolate=true,width=1.791667in,height=0.875000in]{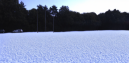}}%
\end{pgfscope}%
\begin{pgfscope}%
\pgfpathrectangle{\pgfqpoint{7.017052in}{4.408846in}}{\pgfqpoint{1.784123in}{0.871154in}}%
\pgfusepath{clip}%
\pgfsetbuttcap%
\pgfsetmiterjoin%
\pgfsetlinewidth{3.011250pt}%
\definecolor{currentstroke}{rgb}{0.913725,0.443137,0.196078}%
\pgfsetstrokecolor{currentstroke}%
\pgfsetdash{}{0pt}%
\pgfpathmoveto{\pgfqpoint{7.017052in}{4.408846in}}%
\pgfpathlineto{\pgfqpoint{8.801175in}{4.408846in}}%
\pgfpathlineto{\pgfqpoint{8.801175in}{5.280000in}}%
\pgfpathlineto{\pgfqpoint{7.017052in}{5.280000in}}%
\pgfpathlineto{\pgfqpoint{7.017052in}{4.408846in}}%
\pgfpathclose%
\pgfusepath{stroke}%
\end{pgfscope}%
\begin{pgfscope}%
\pgfpathrectangle{\pgfqpoint{1.274985in}{3.494135in}}{\pgfqpoint{1.784123in}{0.871154in}}%
\pgfusepath{clip}%
\pgfsys@transformshift{1.274985in}{3.494135in}%
\pgftext[left,bottom]{\includegraphics[interpolate=true,width=1.791667in,height=0.875000in]{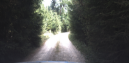}}%
\end{pgfscope}%
\begin{pgfscope}%
\pgfpathrectangle{\pgfqpoint{3.189007in}{3.494135in}}{\pgfqpoint{1.784123in}{0.871154in}}%
\pgfusepath{clip}%
\pgfsys@transformshift{3.189007in}{3.494135in}%
\pgftext[left,bottom]{\includegraphics[interpolate=true,width=1.791667in,height=0.875000in]{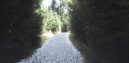}}%
\end{pgfscope}%
\begin{pgfscope}%
\pgfpathrectangle{\pgfqpoint{3.189007in}{3.494135in}}{\pgfqpoint{1.784123in}{0.871154in}}%
\pgfusepath{clip}%
\pgfsetbuttcap%
\pgfsetmiterjoin%
\pgfsetlinewidth{3.011250pt}%
\definecolor{currentstroke}{rgb}{0.913725,0.443137,0.196078}%
\pgfsetstrokecolor{currentstroke}%
\pgfsetdash{}{0pt}%
\pgfpathmoveto{\pgfqpoint{3.189007in}{3.494135in}}%
\pgfpathlineto{\pgfqpoint{4.973130in}{3.494135in}}%
\pgfpathlineto{\pgfqpoint{4.973130in}{4.365288in}}%
\pgfpathlineto{\pgfqpoint{3.189007in}{4.365288in}}%
\pgfpathlineto{\pgfqpoint{3.189007in}{3.494135in}}%
\pgfpathclose%
\pgfusepath{stroke}%
\end{pgfscope}%
\begin{pgfscope}%
\pgfpathrectangle{\pgfqpoint{5.103030in}{3.494135in}}{\pgfqpoint{1.784123in}{0.871154in}}%
\pgfusepath{clip}%
\pgfsys@transformshift{5.103030in}{3.494135in}%
\pgftext[left,bottom]{\includegraphics[interpolate=true,width=1.791667in,height=0.875000in]{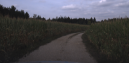}}%
\end{pgfscope}%
\begin{pgfscope}%
\pgfpathrectangle{\pgfqpoint{7.017052in}{3.494135in}}{\pgfqpoint{1.784123in}{0.871154in}}%
\pgfusepath{clip}%
\pgfsys@transformshift{7.017052in}{3.494135in}%
\pgftext[left,bottom]{\includegraphics[interpolate=true,width=1.791667in,height=0.875000in]{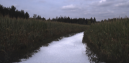}}%
\end{pgfscope}%
\begin{pgfscope}%
\pgfpathrectangle{\pgfqpoint{7.017052in}{3.494135in}}{\pgfqpoint{1.784123in}{0.871154in}}%
\pgfusepath{clip}%
\pgfsetbuttcap%
\pgfsetmiterjoin%
\pgfsetlinewidth{3.011250pt}%
\definecolor{currentstroke}{rgb}{0.913725,0.443137,0.196078}%
\pgfsetstrokecolor{currentstroke}%
\pgfsetdash{}{0pt}%
\pgfpathmoveto{\pgfqpoint{7.017052in}{3.494135in}}%
\pgfpathlineto{\pgfqpoint{8.801175in}{3.494135in}}%
\pgfpathlineto{\pgfqpoint{8.801175in}{4.365288in}}%
\pgfpathlineto{\pgfqpoint{7.017052in}{4.365288in}}%
\pgfpathlineto{\pgfqpoint{7.017052in}{3.494135in}}%
\pgfpathclose%
\pgfusepath{stroke}%
\end{pgfscope}%
\begin{pgfscope}%
\pgfpathrectangle{\pgfqpoint{1.274985in}{2.579423in}}{\pgfqpoint{1.784123in}{0.871154in}}%
\pgfusepath{clip}%
\pgfsys@transformshift{1.274985in}{2.579423in}%
\pgftext[left,bottom]{\includegraphics[interpolate=true,width=1.791667in,height=0.875000in]{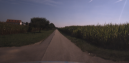}}%
\end{pgfscope}%
\begin{pgfscope}%
\pgfpathrectangle{\pgfqpoint{3.189007in}{2.579423in}}{\pgfqpoint{1.784123in}{0.871154in}}%
\pgfusepath{clip}%
\pgfsys@transformshift{3.189007in}{2.579423in}%
\pgftext[left,bottom]{\includegraphics[interpolate=true,width=1.791667in,height=0.875000in]{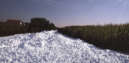}}%
\end{pgfscope}%
\begin{pgfscope}%
\pgfpathrectangle{\pgfqpoint{3.189007in}{2.579423in}}{\pgfqpoint{1.784123in}{0.871154in}}%
\pgfusepath{clip}%
\pgfsetbuttcap%
\pgfsetmiterjoin%
\pgfsetlinewidth{3.011250pt}%
\definecolor{currentstroke}{rgb}{0.913725,0.443137,0.196078}%
\pgfsetstrokecolor{currentstroke}%
\pgfsetdash{}{0pt}%
\pgfpathmoveto{\pgfqpoint{3.189007in}{2.579423in}}%
\pgfpathlineto{\pgfqpoint{4.973130in}{2.579423in}}%
\pgfpathlineto{\pgfqpoint{4.973130in}{3.450577in}}%
\pgfpathlineto{\pgfqpoint{3.189007in}{3.450577in}}%
\pgfpathlineto{\pgfqpoint{3.189007in}{2.579423in}}%
\pgfpathclose%
\pgfusepath{stroke}%
\end{pgfscope}%
\begin{pgfscope}%
\pgfpathrectangle{\pgfqpoint{5.103030in}{2.579423in}}{\pgfqpoint{1.784123in}{0.871154in}}%
\pgfusepath{clip}%
\pgfsys@transformshift{5.103030in}{2.579423in}%
\pgftext[left,bottom]{\includegraphics[interpolate=true,width=1.791667in,height=0.875000in]{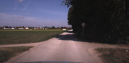}}%
\end{pgfscope}%
\begin{pgfscope}%
\pgfpathrectangle{\pgfqpoint{7.017052in}{2.579423in}}{\pgfqpoint{1.784123in}{0.871154in}}%
\pgfusepath{clip}%
\pgfsys@transformshift{7.017052in}{2.579423in}%
\pgftext[left,bottom]{\includegraphics[interpolate=true,width=1.791667in,height=0.875000in]{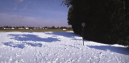}}%
\end{pgfscope}%
\begin{pgfscope}%
\pgfpathrectangle{\pgfqpoint{7.017052in}{2.579423in}}{\pgfqpoint{1.784123in}{0.871154in}}%
\pgfusepath{clip}%
\pgfsetbuttcap%
\pgfsetmiterjoin%
\pgfsetlinewidth{3.011250pt}%
\definecolor{currentstroke}{rgb}{0.913725,0.443137,0.196078}%
\pgfsetstrokecolor{currentstroke}%
\pgfsetdash{}{0pt}%
\pgfpathmoveto{\pgfqpoint{7.017052in}{2.579423in}}%
\pgfpathlineto{\pgfqpoint{8.801175in}{2.579423in}}%
\pgfpathlineto{\pgfqpoint{8.801175in}{3.450577in}}%
\pgfpathlineto{\pgfqpoint{7.017052in}{3.450577in}}%
\pgfpathlineto{\pgfqpoint{7.017052in}{2.579423in}}%
\pgfpathclose%
\pgfusepath{stroke}%
\end{pgfscope}%
\begin{pgfscope}%
\pgfpathrectangle{\pgfqpoint{1.274985in}{1.664712in}}{\pgfqpoint{1.784123in}{0.871154in}}%
\pgfusepath{clip}%
\pgfsys@transformshift{1.274985in}{1.664712in}%
\pgftext[left,bottom]{\includegraphics[interpolate=true,width=1.791667in,height=0.875000in]{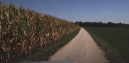}}%
\end{pgfscope}%
\begin{pgfscope}%
\pgfpathrectangle{\pgfqpoint{3.189007in}{1.664712in}}{\pgfqpoint{1.784123in}{0.871154in}}%
\pgfusepath{clip}%
\pgfsys@transformshift{3.189007in}{1.664712in}%
\pgftext[left,bottom]{\includegraphics[interpolate=true,width=1.791667in,height=0.875000in]{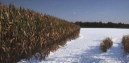}}%
\end{pgfscope}%
\begin{pgfscope}%
\pgfpathrectangle{\pgfqpoint{3.189007in}{1.664712in}}{\pgfqpoint{1.784123in}{0.871154in}}%
\pgfusepath{clip}%
\pgfsetbuttcap%
\pgfsetmiterjoin%
\pgfsetlinewidth{3.011250pt}%
\definecolor{currentstroke}{rgb}{0.913725,0.443137,0.196078}%
\pgfsetstrokecolor{currentstroke}%
\pgfsetdash{}{0pt}%
\pgfpathmoveto{\pgfqpoint{3.189007in}{1.664712in}}%
\pgfpathlineto{\pgfqpoint{4.973130in}{1.664712in}}%
\pgfpathlineto{\pgfqpoint{4.973130in}{2.535865in}}%
\pgfpathlineto{\pgfqpoint{3.189007in}{2.535865in}}%
\pgfpathlineto{\pgfqpoint{3.189007in}{1.664712in}}%
\pgfpathclose%
\pgfusepath{stroke}%
\end{pgfscope}%
\begin{pgfscope}%
\pgfpathrectangle{\pgfqpoint{5.103030in}{1.664712in}}{\pgfqpoint{1.784123in}{0.871154in}}%
\pgfusepath{clip}%
\pgfsys@transformshift{5.103030in}{1.664712in}%
\pgftext[left,bottom]{\includegraphics[interpolate=true,width=1.791667in,height=0.875000in]{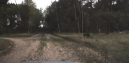}}%
\end{pgfscope}%
\begin{pgfscope}%
\pgfpathrectangle{\pgfqpoint{7.017052in}{1.664712in}}{\pgfqpoint{1.784123in}{0.871154in}}%
\pgfusepath{clip}%
\pgfsys@transformshift{7.017052in}{1.664712in}%
\pgftext[left,bottom]{\includegraphics[interpolate=true,width=1.791667in,height=0.875000in]{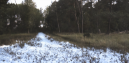}}%
\end{pgfscope}%
\begin{pgfscope}%
\pgfpathrectangle{\pgfqpoint{7.017052in}{1.664712in}}{\pgfqpoint{1.784123in}{0.871154in}}%
\pgfusepath{clip}%
\pgfsetbuttcap%
\pgfsetmiterjoin%
\pgfsetlinewidth{3.011250pt}%
\definecolor{currentstroke}{rgb}{0.913725,0.443137,0.196078}%
\pgfsetstrokecolor{currentstroke}%
\pgfsetdash{}{0pt}%
\pgfpathmoveto{\pgfqpoint{7.017052in}{1.664712in}}%
\pgfpathlineto{\pgfqpoint{8.801175in}{1.664712in}}%
\pgfpathlineto{\pgfqpoint{8.801175in}{2.535865in}}%
\pgfpathlineto{\pgfqpoint{7.017052in}{2.535865in}}%
\pgfpathlineto{\pgfqpoint{7.017052in}{1.664712in}}%
\pgfpathclose%
\pgfusepath{stroke}%
\end{pgfscope}%
\begin{pgfscope}%
\pgfpathrectangle{\pgfqpoint{1.274985in}{0.750000in}}{\pgfqpoint{1.784123in}{0.871154in}}%
\pgfusepath{clip}%
\pgfsys@transformshift{1.274985in}{0.750000in}%
\pgftext[left,bottom]{\includegraphics[interpolate=true,width=1.791667in,height=0.875000in]{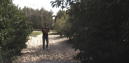}}%
\end{pgfscope}%
\begin{pgfscope}%
\pgfpathrectangle{\pgfqpoint{3.189007in}{0.750000in}}{\pgfqpoint{1.784123in}{0.871154in}}%
\pgfusepath{clip}%
\pgfsys@transformshift{3.189007in}{0.750000in}%
\pgftext[left,bottom]{\includegraphics[interpolate=true,width=1.791667in,height=0.875000in]{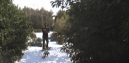}}%
\end{pgfscope}%
\begin{pgfscope}%
\pgfpathrectangle{\pgfqpoint{3.189007in}{0.750000in}}{\pgfqpoint{1.784123in}{0.871154in}}%
\pgfusepath{clip}%
\pgfsetbuttcap%
\pgfsetmiterjoin%
\pgfsetlinewidth{3.011250pt}%
\definecolor{currentstroke}{rgb}{0.913725,0.443137,0.196078}%
\pgfsetstrokecolor{currentstroke}%
\pgfsetdash{}{0pt}%
\pgfpathmoveto{\pgfqpoint{3.189007in}{0.750000in}}%
\pgfpathlineto{\pgfqpoint{4.973130in}{0.750000in}}%
\pgfpathlineto{\pgfqpoint{4.973130in}{1.621154in}}%
\pgfpathlineto{\pgfqpoint{3.189007in}{1.621154in}}%
\pgfpathlineto{\pgfqpoint{3.189007in}{0.750000in}}%
\pgfpathclose%
\pgfusepath{stroke}%
\end{pgfscope}%
\begin{pgfscope}%
\pgfpathrectangle{\pgfqpoint{5.103030in}{0.750000in}}{\pgfqpoint{1.784123in}{0.871154in}}%
\pgfusepath{clip}%
\pgfsys@transformshift{5.103030in}{0.750000in}%
\pgftext[left,bottom]{\includegraphics[interpolate=true,width=1.791667in,height=0.875000in]{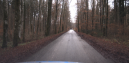}}%
\end{pgfscope}%
\begin{pgfscope}%
\pgfpathrectangle{\pgfqpoint{7.017052in}{0.750000in}}{\pgfqpoint{1.784123in}{0.871154in}}%
\pgfusepath{clip}%
\pgfsys@transformshift{7.017052in}{0.750000in}%
\pgftext[left,bottom]{\includegraphics[interpolate=true,width=1.791667in,height=0.875000in]{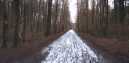}}%
\end{pgfscope}%
\begin{pgfscope}%
\pgfpathrectangle{\pgfqpoint{7.017052in}{0.750000in}}{\pgfqpoint{1.784123in}{0.871154in}}%
\pgfusepath{clip}%
\pgfsetbuttcap%
\pgfsetmiterjoin%
\pgfsetlinewidth{3.011250pt}%
\definecolor{currentstroke}{rgb}{0.913725,0.443137,0.196078}%
\pgfsetstrokecolor{currentstroke}%
\pgfsetdash{}{0pt}%
\pgfpathmoveto{\pgfqpoint{7.017052in}{0.750000in}}%
\pgfpathlineto{\pgfqpoint{8.801175in}{0.750000in}}%
\pgfpathlineto{\pgfqpoint{8.801175in}{1.621154in}}%
\pgfpathlineto{\pgfqpoint{7.017052in}{1.621154in}}%
\pgfpathlineto{\pgfqpoint{7.017052in}{0.750000in}}%
\pgfpathclose%
\pgfusepath{stroke}%
\end{pgfscope}%
\end{pgfpicture}%
\makeatother%
\endgroup%

%% file: tex/goose_colors.tex
\definecolor{undefined}{RGB}{0,0,0}
\definecolor{traffic_cone}{RGB}{255,255,0}
\definecolor{snow}{RGB}{209,87,160}
\definecolor{cobble}{RGB}{255,52,255}
\definecolor{obstacle}{RGB}{255,74,70}
\definecolor{leaves}{RGB}{0,137,65}
\definecolor{street_light}{RGB}{0,111,166}
\definecolor{bikeway}{RGB}{163,0,89}
\definecolor{ego_vehicle}{RGB}{255,219,229}
\definecolor{pedestrian_crossing}{RGB}{122,73,0}
\definecolor{road_block}{RGB}{0,0,166}
\definecolor{road_marking}{RGB}{99,255,172}
\definecolor{car}{RGB}{183,151,98}
\definecolor{bicycle}{RGB}{0,77,67}
\definecolor{person}{RGB}{143,176,255}
\definecolor{bus}{RGB}{153,125,135}
\definecolor{forest}{RGB}{90,0,7}
\definecolor{bush}{RGB}{128,150,147}
\definecolor{traffic_light}{RGB}{27,68,0}
\definecolor{motorcycle}{RGB}{79,198,1}
\definecolor{sidewalk}{RGB}{59,93,255}
\definecolor{curb}{RGB}{74,59,83}
\definecolor{asphalt}{RGB}{255,47,128}
\definecolor{gravel}{RGB}{97,97,90}
\definecolor{boom_barrier}{RGB}{52,54,45}
\definecolor{rail_track}{RGB}{107,121,0}
\definecolor{tree_crown}{RGB}{0,194,160}
\definecolor{tree_root}{RGB}{196,164,132}
\definecolor{tree_trunk}{RGB}{255,170,146}
\definecolor{debris}{RGB}{136,111,76}
\definecolor{crops}{RGB}{0,134,237}
\definecolor{soil}{RGB}{209,97,0}
\definecolor{rider}{RGB}{221,239,255}
\definecolor{animal}{RGB}{0,0,53}
\definecolor{truck}{RGB}{123,79,75}
\definecolor{on_rails}{RGB}{161,194,153}
\definecolor{caravan}{RGB}{48,0,24}
\definecolor{trailer}{RGB}{10,166,216}
\definecolor{building}{RGB}{1,51,73}
\definecolor{wall}{RGB}{0,132,111}
\definecolor{rock}{RGB}{55,33,1}
\definecolor{fence}{RGB}{255,181,0}
\definecolor{guard_rail}{RGB}{194,255,237}
\definecolor{bridge}{RGB}{160,121,191}
\definecolor{tunnel}{RGB}{204,7,68}
\definecolor{pole}{RGB}{192,185,178}
\definecolor{traffic_sign}{RGB}{194,255,153}
\definecolor{misc_sign}{RGB}{0,30,9}
\definecolor{barrier_tape}{RGB}{190,196,89}
\definecolor{kick_scooter}{RGB}{111,0,98}
\definecolor{low_grass}{RGB}{12,189,102}
\definecolor{high_grass}{RGB}{238,195,255}
\definecolor{scenery_vegetation}{RGB}{69,109,117}
\definecolor{sky}{RGB}{183,123,104}
\definecolor{water}{RGB}{122,135,161}
\definecolor{wire}{RGB}{255,140,0}
\definecolor{outlier}{RGB}{120,141,102}
\definecolor{heavy_machinery}{RGB}{250,208,159}
\definecolor{container}{RGB}{255,138,154}
\definecolor{hedge}{RGB}{232, 211, 23}
\definecolor{moss}{RGB}{180, 168, 189}
\definecolor{barrel}{RGB}{208, 208, 0}
\definecolor{pipe}{RGB}{221, 0, 0}
\definecolor{military_vehicle}{RGB}{64,64,64}